\documentclass{article}

 \usepackage[preprint]{neurips_2026}

\usepackage[utf8]{inputenc} 
\usepackage[T1]{fontenc}    
\usepackage{hyperref}       
\usepackage{url}            
\usepackage{booktabs}       
\usepackage{amsfonts}       
\usepackage{nicefrac}       
\usepackage{microtype}      
\usepackage{xcolor}         

\usepackage[most]{tcolorbox}

\usepackage{amsmath}
\usepackage{amssymb}
\usepackage{mathtools}
\usepackage{amsthm}
\usepackage{multirow}
\usepackage{wrapfig}  
\usepackage[capitalize,noabbrev]{cleveref}

\theoremstyle{plain}

\theoremstyle{definition}

\theoremstyle{remark}

\usepackage{enumitem}
\usepackage{pifont}

\usepackage[table]{xcolor} 

\newcommand{\yichen}[1]{{\color{black} #1}}
\newcommand{\kailong}[1]{{\color{black} #1}}

\usepackage[textsize=tiny]{todonotes}
\title{Formatting Instructions For NeurIPS 2026}

\newcommand*\samethanks[1][\value{footnote}]{\footnotemark[#1]}
\author{%
  Yichen Wu\textsuperscript{1}\footnotemark[1]\thanks{Equal Contribution}, ~ Kailong Fan\textsuperscript{1}\samethanks, ~ Sangjoon Park\textsuperscript{2}\samethanks, ~Yuhan Liu\textsuperscript{3},~Zhiyi Shi\textsuperscript{1},\\
  {\bfseries~Sekeun Kim\textsuperscript{1},~Dania Daye\textsuperscript{4}, ~Hana Farzaneh\textsuperscript{1}, ~Wei Liu\textsuperscript{5}, ~Xiang Li\textsuperscript{1}\thanks{Corresponding authors: Yujin Oh(yujinoh@yuhs.ac), Xiang Li and Quanzheng Li~(\{xli60,quanzheng\}@mgh.harvard.edu)}, }\\ 
  {\bfseries~Raul Uppot\textsuperscript{1},~Yujin Oh\textsuperscript{1,2}\samethanks, ~Quanzheng Li\textsuperscript{1}\samethanks }\\
  \textsuperscript{1}Harvard Medical School/MGH, \textsuperscript{2}Yonsei University, \textsuperscript{3}MBZUAI, \textsuperscript{4}UW-Madison, \textsuperscript{5}Mayo Clinic \\
  \{yiwu6, kfan5\}@mgh.harvard.edu, 
}

\begin{document}

\title{CoMMa: Contribution-Aware Medical Multi-Agents for Decentralized Oncology Decision Support}
\maketitle

\begin{abstract}
\yichen{Recent multi-agent frameworks have shown promise for oncology decision support, yet most assume centralized data access and rely on prompt-based assignment, limiting their applicability in privacy-sensitive clinical settings. We propose Contribution-Aware Medical Multi-Agents (CoMMa), a decentralized LLM-agent framework where specialists operate on partitioned clinical data streams. Unlike prior approaches that share inputs across agents, CoMMa enforces data decentralization to include stronger role specialization and further enhances this via agent-specific finetuning. To enable reliable and interpretable coordination, we introduce a contribution-aware aggregation mechanism that replaces stochastic, narrative-based reasoning with deterministic embedding projections to approximate each agent's marginal utility. This yields explicit credit assignment over agents, providing a stable and interpretable decision pathway aligned with clinical requirements. We evaluate CoMMa on multiple oncology benchmarks, including real-world multidisciplinary tumor board datasets from large academic hospitals in North America and East Asia, as well as public datasets, demonstrating strong performance and generalization across heterogeneous clinical settings.}
  
\end{abstract}

\section{Introduction}
Effective clinical decision support in oncology, such as treatment recommendation, prognosis estimation, and recurrence prediction, requires the seamless integration of longitudinal, heterogeneous data. {In practice, this process is carried out through multidisciplinary tumor boards~\cite{taylor2010multidisciplinary, patkar2011cancer}, where clinicians jointly} 
{discuss}
patient histories, medical imaging, pathology findings, and evolving laboratory markers. 
Recently, large language model (LLM)-based multi-agents systems have emerged as a promising paradigm for this setting, \yichen{distributing complex reasoning across specialized agents to better reflect multidisciplinary workflows~\cite{chen2025enhancing, kim2024mdagents, ferber2025development, han2025multi, fallahpour2025medrax}.}

\begin{figure}[t]
  \centering
  \includegraphics[width=1\linewidth]{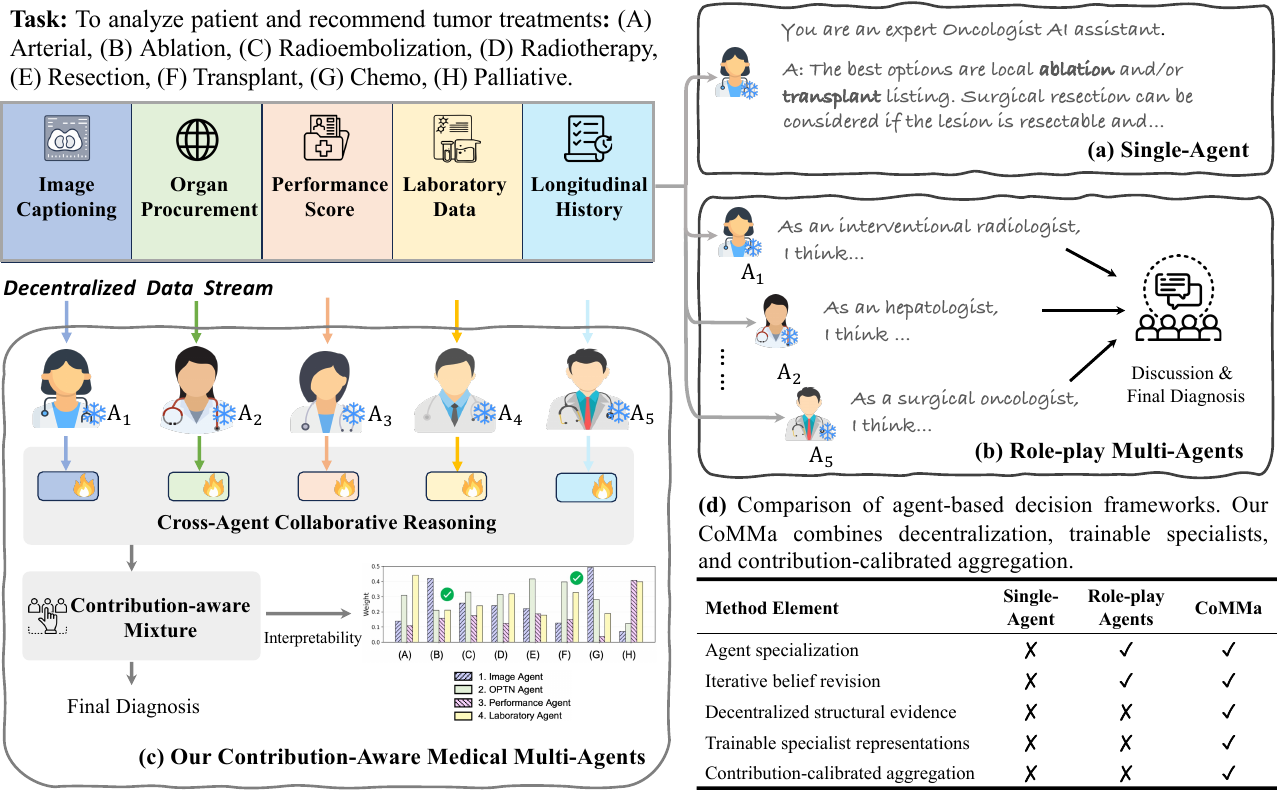}
  \caption{\yichen{Comparison of medical decision-making frameworks. (a) Single-agent and (b) role-based multi-agent systems rely on shared inputs and narrative reasoning. (c) CoMMa introduces decentralized data streams, agent-specific specialization, and contribution-aware aggregation for interpretable and stable clinical decision-making. (d) Summary table comparing key design elements across the three decision-making frameworks under a unified formulation. }}
   \label{fig:main}
\end{figure}

Despite their promise, \yichen{most existing medical multi-agent frameworks rely on role-based prompting, where agents are assigned clinician-like roles such as oncologists or radiologists~\cite{chen2025enhancing, kim2024mdagents, zhang2025multi}. As shown in Fig.~\ref{fig:main}(b), these agents typically share the same full patient information and communicate through natural language discussion. However, because all agents access the same full patient context, their specialization is mainly defined by prompts rather than distinct clinical evidence or domain-specific adaptation. As a result, agents with different clinical roles often rely on overlapping evidence, making it difficult to determine which information or which agent contributes most to the final decision, even though such interpretability is particularly important in clinical decision-making. Compared with the single-agent setting in Fig.~\ref{fig:main}(a), current multi-agent frameworks in Fig.~\ref{fig:main}(b) provide improved collaboration, but still depend on shared patient context, limited structural specialization, and weak attribution of agent contributions in complex, real-world clinical decision workflows.}

To address these limitations, we propose Contribution-Aware Medical Multi-Agents (CoMMa), \yichen{a framework that moves beyond prompt-based role assignment toward data-decentralized and contribution-aware specialization. As illustrated in Fig.~\ref{fig:main}(c), CoMMa partitions the clinical context into distinct evidence streams and assigns each stream to a dedicated agent, enabling specialization through separate clinical evidence rather than persona simulation alone. Each agent further uses a domain-specific adaptation module to strengthen specialty-aware representations. Agents then exchange information through deterministic representation-level collaboration, producing updated agent states without relying on free-form natural language discussion.}

\yichen{To support interpretable clinical decision-making, the proposed CoMMa introduces a contribution-aware aggregation mechanism that explicitly models the relative contribution of each specialist agent to the final prediction. Inspired by cooperative game theory, CoMMa regularizes agent contributions using Shapley-based marginal utility estimation~\cite{shapley1953value, beechey2023icml}, enabling more transparent and structured multi-agent reasoning. This design enables structurally specialized collaboration and explicit agent-level attribution without requiring full-context sharing across agents, as summarized in Fig.~\ref{fig:main} (c)(d). We evaluate CoMMa across diverse oncology benchmarks, including real-world multidisciplinary tumor board datasets from large academic hospitals in North America and East Asia, together with public benchmarks. Experimental results demonstrate that CoMMa consistently improves predictive performance, decision stability, and attribution interpretability across multiple open-source LLM backbones and clinical decision-making tasks.}

{In summary, our core contributions are summarized as follows: }
\begin{itemize}[leftmargin=3.5mm, itemsep=0mm, topsep=-0.5mm]

\item We propose CoMMa, a decentralized medical multi-agent framework that enables structurally specialized reasoning without requiring full-context sharing across agents.

\item We introduce a Shapley-regularized contribution-aware aggregation strategy, enabling explicit attribution for interpretable clinical decision-making.

\item We validate CoMMa across multiple open-source LLM backbones and diverse real-world oncology benchmarks from large academic hospitals in North America and East Asia, demonstrating consistent improvements over existing single-agent and multi-agent baselines.

\end{itemize}

\enlargethispage{\baselineskip}

\section{Related Works}
\textbf{Multi-Agents for Medical Applications.} Recent studies have explored multi-agent LLM systems {for collaborative clinical decision-making, where agents take specialized medical roles~\cite{chen2025evaluating,tang2025medagentsbench,zou2025rise,johri2025evaluation,wang2025medagent,zhou2025mam,almansoori2025medagentsim,fallahpour2025medrax}.}  
{MedAgents~\cite{tang2024medagents} frames problem-solving as a multi-physician debate with answer aggregation, while MDAgents~\cite{kim2024mdagents} adaptively chooses single-versus multi-agent collaboration to balance accuracy and cost.} Beyond free-form discussion, Multi-Agent Conversation~\cite{chen2025enhancing} structures deliberation to mimic multidisciplinary meetings, {and AI Hospital~\cite{fan2025ai} evaluates longitudinal reasoning in a multi-agent clinical simulation. However, these systems are primarily role-based: collaboration is mediated through narrative exchanges and the final decision is fused implicitly, making evidence-level contribution hard to quantify. In contrast, our contribution-aware framework explicitly attributes how each evidence source and agent influences the final prediction, enabling more controlled fusion and more reliable clinical decision outputs.}

\textbf{Game Theory in Clinical Decision-Making.} Game theory provides mathematical frameworks for modeling strategic interactions among rational agents and has been widely applied in general-domain decision-making, including economics, management, defense, and computer science \cite{von2007theory, dutang2013competition, ho2022game, parsons2002game, lanctot2017neurips}. Despite its strong theoretical foundations, its adoption in clinical settings remains limited \cite{mendoncca2020improving, lau2023game}, particularly in oncology. Oncology decision-making relies on heterogeneous and interdependent evidence sources, where quantifying the contribution of each component is essential for interpretability and clinical reliability. Among cooperative game-theoretic approaches, the Shapley value \cite{shapley1953value, beechey2023icml} provides a principled and axiomatized solution for fair allocation of a total payoff among participating agents. In a clinical AI setting, the payoff corresponds to the final clinical decision or predictive outcome, while each agent represents a distinct modality or specialized source of clinical information. The Shapley value of an agent is defined as its expected marginal contribution, averaged over all possible coalitions, satisfying desirable payoff. To the best of our knowledge, this work is the first to incorporate Shapley value–based contribution modeling as a regularization mechanism in a multi-agent oncological tumor board framework.

\section{Method}
\label{sec:method}
\yichen{To enable structurally specialized and interpretable clinical decision-making, we propose CoMMa, a contribution-aware multi-agent framework for clinical outcome prediction.}
Given heterogeneous patient information $\mathbf{x}=\{\mathbf{x}_1,\ldots,\mathbf{x}_N\}$, each $\mathbf{x}_i$ \yichen{denotes a distinct clinical evidence stream, such as} imaging findings, laboratory measurements, staging information, or clinical history.
\yichen{Instead of concatenating all streams into a single input, CoMMa keeps them separated and assigns each stream to a specialist agent. The framework then learns how much each agent should contribute to each prediction for each patient. The design preserves the complementary structure of clinical evidence and provides explicit agent-level attribution for the final decision. }

\begin{figure*}[!ht]
  \centering
  \includegraphics[width=0.85\linewidth]{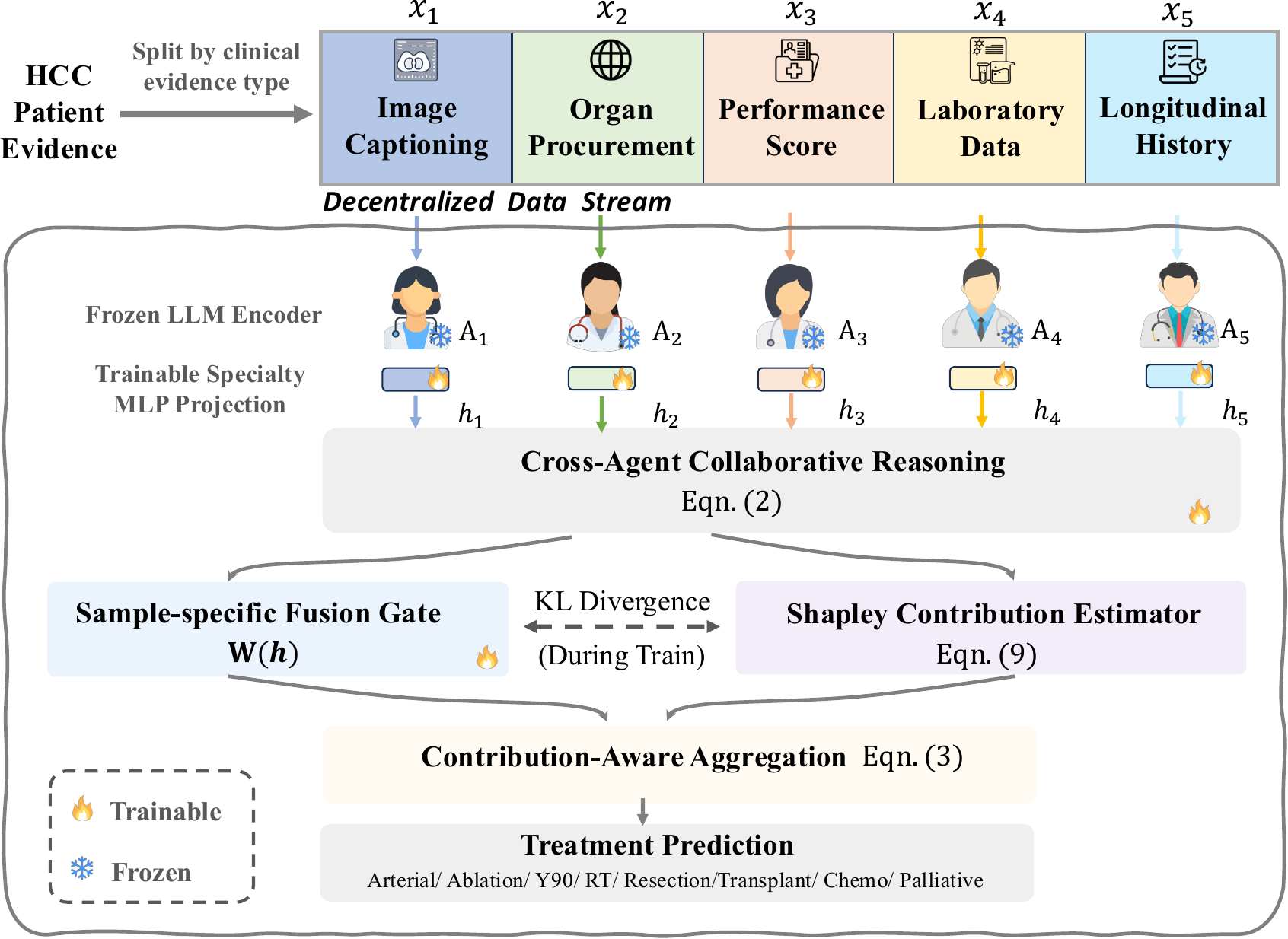}
  \caption{Architectural overview of CoMMa for decentralized multi-agent oncology decision support. Patient evidence is partitioned into distinct clinical streams and processed by specialist agents with separate observation spaces. Agent-specific representations are refined through cross-agent collaborative reasoning and aggregated by a sample-specific contribution gate regularized with Shapley-based attribution during model training.} 
   \label{fig:method}
\vspace{-2mm}
\end{figure*}

\yichen{As illustrated in Fig.~\ref{fig:method}, patient evidence is first partitioned into decentralized clinical streams and processed by specialist agents. Each stream is encoded using a frozen open-source LLM encoder followed by a lightweight trainable specialty projection module, producing agent-specific representations. These representations are then refined through cross-agent collaborative reasoning, allowing specialist agents to exchange complementary information while maintaining their evidence-specific roles. Finally, CoMMa performs contribution-aware aggregation through a sample-specific fusion gate regularized by a Shapley contribution estimator, generating the final clinical prediction with explicit attribution of agent contributions.}

\subsection{Decentralized Specialist Encoding}
\label{sec:encoding}
\yichen{}
CoMMa constructs each specialist agent with restricted access to one clinical evidence stream. The $i$-th agent observes only $\mathbf{x}_i$, rather than the full patient context, so its role is defined by its observation space instead of textual role prompting. Concretely, each stream is encoded by a frozen LLM backbone and an agent-specific trainable projection head:
\begin{equation}
\mathbf{e}_i =
\mathrm{LLM}_{\mathrm{frozen}}(\mathbf{x}_i)[\texttt{<CLS>}],
\qquad
\mathbf{h}_i =
f_{\theta_i}(\mathbf{e}_i),
\quad i=1,\ldots,N,
\end{equation}
where $\mathbf{e}_i\in\mathbb{R}^{d}$ is the deterministic representation of the $i$-th evidence stream, and $\mathbf{h}_i\in\mathbb{R}^{C}$ denotes the agent output over $C$ prediction targets. The projection head $f_{\theta_i}$ is optimized only for its assigned stream, enabling agent-specific adaptation while keeping the LLM backbone fixed.

\yichen{This decentralized design allows specialist agents to maintain distinct observation spaces and representation pathways throughout collaborative decision-making. In practice, frozen LLM embeddings are cached and reused, so training focuses on the lightweight agent-specific modules.}

\subsection{Cross-Agent Collaborative Reasoning}
\label{sec:collaboration}
\yichen{Although each specialist agent observes only one clinical evidence stream, effective clinical decision-making still requires integrating complementary information across agents. CoMMa therefore performs collaboration only after decentralized encoding, where agents exchange compact latent representations instead of accessing each other’s raw clinical evidence or the full patient context.} 

\yichen{For each agent, we summarize the peer representations by averaging the outputs of all other agents, and the $i$-th agent then refines its representation through a gated residual update: }

\begin{equation}
\bar{\mathbf{h}}_{-i}
=
\frac{1}{N-1}\sum_{j\neq i}\mathbf{h}_j,
\qquad
\hat{\mathbf{h}}_i
=
\mathbf{h}_i
+
\rho\,
\sigma\!\left(g_{\psi}([\mathbf{h}_i,\bar{\mathbf{h}}_{-i}])\right)
\odot
u_{\psi}([\mathbf{h}_i,\bar{\mathbf{h}}_{-i}]).
\end{equation}
Here, $g_{\psi}$ and $u_{\psi}$ are lightweight trainable networks, $\sigma(\cdot)$ denotes the sigmoid function, $\odot$ denotes element-wise multiplication, and $\rho$ controls the residual update scale. This interaction enables each agent to incorporate complementary peer signals while preserving its own evidence-specific representation throughout collaborative clinical decision-making. Unlike natural language debate, the collaboration is deterministic and operates directly on compact agent-level representations.

\subsection{Contribution-Aware Aggregation and Learning}
\label{sec:contribution}

After cross-agent collaborative reasoning, CoMMa aggregates the refined specialist outputs through a sample-specific contribution gate. For each patient and prediction target, the gate assigns a normalized weight to each agent based on its refined representation:
\begin{equation}
W_{i,k}
=
\frac{
\exp\left(G_{\phi}(\hat{\mathbf{h}}_i,\mathbf{H})_{k}/\tau\right)
}{
\sum_{j=1}^{N}
\exp\left(G_{\phi}(\hat{\mathbf{h}}_j,\mathbf{H})_{k}/\tau\right)
},
\qquad
\hat{y}_k
=
\sum_{i=1}^{N} W_{i,k}\hat{h}_{i,k},
\end{equation}
where $\mathbf{H}=\{\hat{\mathbf{h}}_1,\ldots,\hat{\mathbf{h}}_N\}$, $G_{\phi}$ is a lightweight contribution gate, $\tau$ is a temperature parameter, and $\hat{y}_k$ is the prediction for target $k$. The matrix $\mathbf{W}$ provides a target-specific decomposition of the final prediction across specialist agents.

Learning meaningful contribution weights requires more than prediction supervision alone. CoMMa therefore optimizes three complementary objectives:
\begin{equation}
\mathcal{L}_{\mathrm{total}}
=
\mathcal{L}_{\mathrm{BCE}}(\hat{\mathbf{y}},\mathbf{y})
+
\lambda_{\mathrm{pg}}\mathcal{L}_{\mathrm{pg}}
+
\lambda_{\mathrm{shap}}\mathcal{L}_{\mathrm{shap}} .
\end{equation}

\begin{tcolorbox}[colback=gray!5,colframe=gray!40,boxrule=0.3pt,left=3pt,right=3pt,top=3pt,bottom=3pt]
\textbf{Objective.}
$\mathcal{L}_{\mathrm{BCE}}$ optimizes the final clinical prediction;
$\mathcal{L}_{\mathrm{pg}}$ encourages agents with higher marginal utility to receive larger contribution weights;
$\mathcal{L}_{\mathrm{shap}}$ regularizes the learned weights using Shapley-based coalition utility, enabling more stable and interpretable contribution attribution.
\end{tcolorbox}

\paragraph{Utility-guided contribution learning.}
To align the contribution gate with each agent's actual predictive utility, CoMMa first estimates how much each specialist agent contributes to the final prediction. We adopt a leave-one-out strategy that measures the prediction loss change after removing one agent from the aggregation process. Let $\bar{\mathbf{y}}$ denote the aggregated prediction using all refined agents, and let $\bar{\mathbf{y}}_{-i}$ denote the corresponding prediction after excluding agent $i$. The marginal utility of agent $i$ for target $k$ is defined as
\begin{equation}
r_{i,k}
=
\mathcal{L}^{(k)}_{\mathrm{BCE}}(\bar{\mathbf{y}}_{-i},\mathbf{y})
-
\mathcal{L}^{(k)}_{\mathrm{BCE}}(\bar{\mathbf{y}},\mathbf{y}) .
\end{equation}

A positive value indicates that removing agent $i$ increases the prediction loss, suggesting that the agent provides useful evidence for target $k$. Since the raw utility scores may vary across agents and prediction targets, we further center the rewards across agents to obtain an advantage signal:
\begin{equation}
A_{i,k}
=
r_{i,k}
-
\frac{1}{N}\sum_{j=1}^{N} r_{j,k}.
\end{equation}

This advantage signal serves as utility-based supervision for the contribution gate, encouraging the learned aggregation weights to align with each agent's estimated marginal utility:
\begin{equation}
\mathcal{L}_{\mathrm{pg}}
=
-
\mathbb{E}
\left[
\sum_{i=1}^{N}
\sum_{k=1}^{C}
A_{i,k}\log W_{i,k}
\right].
\end{equation}

This objective encourages the contribution gate to consistently assign larger weights to agents that improve predictive performance.

\paragraph{Shapley-based contribution regularization.}
Although leave-one-out rewards provide a direct estimate of agent utility, they only evaluate contribution under a single exclusion setting and can therefore be unstable in heterogeneous clinical scenarios. To obtain a more robust contribution target, we further adopt the Shapley principle from cooperative game theory~\cite{shapley1953value, beechey2023icml}, which estimates each agent's average marginal contribution across different collaborating coalitions.

For each agent and prediction target, the Shapley target $\Phi_{i,k}$ is computed as,
\begin{equation}
\Phi_{i,k}
\propto
\mathbb{E}_{S\subseteq\mathcal{N}\setminus\{i\}}
\left[
\mathcal{L}^{(k)}_{\mathrm{BCE}}(S,\mathbf{y})
-
\mathcal{L}^{(k)}_{\mathrm{BCE}}(S\cup\{i\},\mathbf{y})
\right]_{+},
\qquad
\sum_{i=1}^{N}\Phi_{i,k}=1 .
\end{equation}

Intuitively, this formulation measures how much agent $i$ improves prediction performance when added to different subsets of collaborating agents, providing a coalition-aware estimate of contribution importance beyond pairwise or leave-one-out comparisons. The learned contribution gate is then regularized toward this Shapley target:
\begin{equation}
\mathcal{L}_{\mathrm{shap}}
=
\mathbb{E}
\left[
\mathrm{KL}\!\left(\boldsymbol{\Phi}\,\|\,\mathbf{W}\right)
\right].
\end{equation}

\yichen{\textbf{Remark.} We use the term multi-agent in a structural sense: each specialist has a restricted observation space, an agent-specific transformation, a latent state, and an explicit update conditioned on other agents' states. Unlike standard multi-view fusion, CoMMa refines evidence-specific agent states through inter-agent communication before assigning target-specific contributions to the final clinical decision in heterogeneous clinical decision-making with decentralized information sources.}

\section{Experiments}
\label{sec:res}

\subsection{Experimental Settings}

\textbf{Datasets and Implementation Details.}
\yichen{Experiments are conducted across four oncology benchmarks:}
\kailong{a real-world multidisciplinary tumor board dataset from a large academic healthcare system in North America (HCC Tumorboard; 116 patients), two public oncology benchmarks (MTBBench; 66
patient cases and 573 expert-validated question-answer pairs~\cite{vasilev2025mtbbench}, and LungCure; 999 patients~\cite{hao2026lungcure}), and an externally annotated clinical dataset from an
academic hospital system in East Asia for cross-domain evaluation and interpretability analysis, which we refer to as \textsc{TumorBoardRank} (93 tumor board cases and 276 pairwise option comparisons after filtering single-option cases).}  
\yichen{Additional details on clinical variables, evidence partitioning, preprocessing, and cohort statistics are provided in Appendix~\cref{app:data}.}
CoMMa is implemented using LangGraph~\cite{langchain} and evaluated with multiple LLM backbones, including Llama3.1-70B-Instruct~\cite{grattafiori2024llama}, Qwen2.5-72B~\cite{qwen2_5}, and Gemma4-
31B~\cite{google_gemma4_31b_it}. Unless otherwise specified, experiments use four specialist agents and one round of cross-agent collaborative reasoning with residual scale $\rho=0.5$;
\textsc{TumorBoardRank} uses three specialist agents corresponding to basic profile, clinical tests, and treatment history. HCC experiments use learning rate $2\times10^{-5}$, 121 epochs, $
\lambda_{\mathrm{pg}}=0.07$, and $\lambda_{\mathrm{shap}}=1.0$. For MTBBench, we use learning rate $10^{-5}$, 31 epochs, $\lambda_{\mathrm{pg}}=1$, and $\lambda_{\mathrm{shap}}=1.0$. For LungCure,
we use learning rate $5\times10^{-5}$, 100 epochs, $\lambda_{\mathrm{pg}}=0.1$, and $\lambda_{\mathrm{shap}}=2.0$. For \textsc{TumorBoardRank}, we use Gemma4-31B with learning rate $5\times10^{-5}$,
100 epochs, $\lambda_{\mathrm{pg}}=0.1$, and $\lambda_{\mathrm{shap}}=4.0$.
We use two NVIDIA A100 80GB GPUs for training each model.


\textbf{Evaluation Protocol.} We report Area Under the ROC Curve (AUC) and Accuracy at a fixed threshold of 0.5. For HCC, AUC and Acc are first computed for each treatment category and then macro-averaged across valid categories. If a validation split contains only one label value for a category, that category is excluded from the HCC macro-AUC calculation. For MTBBench, the progression and recurrence endpoints are evaluated as separate binary tasks.
\kailong{For LungCure, we evaluate six binary staging indicators spanning primary tumor extent (T), regional nodal involvement (N), and distant metastasis (M), and report macro-averaged AUC and Acc across these indicators.
For \textsc{TumorBoardRank}, we evaluate candidate management-option ranking using Top-1 Acc, Case Distribution Acc, and Pair Order Acc.
Top-1 Acc measures whether the highest-ranked predicted option matches the option with the highest physician consensus; Case Distribution Acc measures whether the full predicted option ordering matches the physician-consensus ordering; and Pair Order Acc measures the fraction of option pairs ordered consistently with physician consensus.
}
All reported summary results are computed over three random seeds, using seeds 1000, 2000, and 3000.


\textbf{Baseline Methods.}
We compare CoMMA with single-agent and multi-agent LLM baselines under the same evaluation protocol. The direct single-agent baseline receives the full patient case as one input and outputs a probability for each target without explicit multi-agent interaction. MDAgents Basic \cite{kim2024mdagents} is implemented as an adapted medical decision-agent baseline. For HCC, we use the adapted local-LLM MDAgents script with the same HCC label space; for MTBBench, we use the corresponding binary prognosis version. MAC \cite{chen2025enhancing} is implemented as a multi-agent consultation baseline in which multiple role-specific doctor agents iteratively estimate the target probability, and the final prediction is obtained from the last consensus round. AI-Hospital~\cite{fan2025ai} is implemented as a collaborative consultation baseline in which several doctor agents provide specialty estimates and a host physician aggregates them into a final probability.

\subsection{Main Results}
\textbf{Comparison on different oncology benchmarks.}
Table~\ref{tab:res} compares CoMMa with single-agent and existing multi-agent baselines across HCC Tumorboard, Lung Cure, and MTBBench. CoMMa consistently achieves the best overall performance across benchmarks and LLM backbones. On HCC Tumorboard, CoMMa obtains the highest treatment recommendation accuracy under all three backbones, improving Llama3.1-70B from 0.690 to 0.795 and Qwen2.5-72B from 0.650 to 0.750 over the single-agent baseline. On Lung Cure, CoMMa also improves TNM stage prediction, achieving the best accuracy with Qwen2.5-72B (0.802) and Gemma-31B (0.830). The largest gains appear on MTBBench, where CoMMa substantially improves progression and recurrence prediction. 
These results show that CoMMa generalizes across real-world and public oncology benchmarks while consistently improving predictive performance.

\textbf{Comparison on various LLM backbones.}
CoMMa shows consistent gains across Llama3.1-70B, Qwen2.5-72B, and Gemma-31B, indicating that the framework is not tied to a specific backbone. With Llama3.1-70B, CoMMa improves HCC treatment recommendation AUC from 0.590 to 0.711 and accuracy from 0.690 to 0.795 over the single-agent baseline. It also improves MTBBench progression AUC from 0.503 to 0.756 and recurrence AUC from 0.526 to 0.711. Similar trends are observed with Qwen2.5-72B, where CoMMa achieves the best results among open-source baselines across all tasks, including Lung Cure accuracy of 0.802 and MTBBench recurrence AUC of 0.845. With Gemma-31B, CoMMa yields especially large improvements on MTBBench, increasing progression AUC from 0.556 to 0.905 and recurrence AUC from 0.534 to 0.911 over the single-agent baseline. Overall, these results show that decentralized specialist encoding and contribution-aware aggregation provide robust improvements across different LLM families and model scales.

\begin{table*}[t]
\caption{Performance comparison on three oncology benchmarks: HCC Tumorboard, Lung Cure, and MTBBench. CoMMa consistently improves predictive performance over single-agent and existing multi-agent baselines across multiple open-source LLM backbones. }
\resizebox{1\linewidth}{!}{
\label{tab:res}
\centering
\begin{tabular}{lccccccccc}

\toprule
\multirow{3}{*}{\bf{Methods}} 
& \multicolumn{2}{c}{\bf{HCC Tumorboard}} 
& \multicolumn{2}{c}{\bf{Lung Cure}} 
& \multicolumn{4}{c}{{\bf{MTBBench}}}\\
\cmidrule(r){2-3} \cmidrule(r){4-5} \cmidrule(r){6-9}

& \multicolumn{2}{c}{{Treatment Recommendation}} 
& \multicolumn{2}{c}{{TNM Stage Prediction}} 
& \multicolumn{2}{c}{{Progression}} 
& \multicolumn{2}{c}{{Recurrence}} \\
\cmidrule(r){2-3} \cmidrule(r){4-5} \cmidrule(r){6-7} \cmidrule(r){8-9}

& {AUC} & {Acc} 
& {AUC} & {Acc} 
& {AUC} & {Acc} 
& {AUC} & {Acc} \\
\cmidrule(r){1-1} \cmidrule(r){2-3} \cmidrule(r){4-5} \cmidrule(r){6-7} \cmidrule(r){8-9}

\multicolumn{9}{l}{\textit{\small GPT 4.1}} \\
\multirow{1}{*}{In-context learning}
& N/A & 0.784 $\pm$ 0.007
& N/A &  0.798 $\pm$0.070  
& N/A & 0.485 $\pm$ 0.113 
& N/A & 0.425 $\pm$ 0.086 \\

\cmidrule(r){1-1} \cmidrule(r){2-3} \cmidrule(r){4-5} \cmidrule(r){6-7} \cmidrule(r){8-9}

\multicolumn{9}{l}{\textit{\small Llama3.1-70B~\cite{grattafiori2024llama}}} \\
\multirow{1}{*}{Single-Agent}
& 0.590 $\pm$ 0.066 & 0.690 $\pm$ 0.064
& 0.782 $\pm$ 0.061 & 0.735 $\pm$ 0.044 
& 0.503 $\pm$ 0.001 & 0.519 $\pm$ 0.004 
& 0.526 $\pm$ 0.018 & 0.491 $\pm$ 0.011 
 \\

\multirow{1}{*}{MDAgents~\cite{kim2024mdagents}}
& 0.693 $\pm$ 0.040 & 0.708 $\pm$ 0.019
& 0.781 $\pm$ 0.051 & 0.746 $\pm$ 0.048 
& 0.523 $\pm$ 0.010 & 0.526 $\pm$ 0.032
& 0.536 $\pm$ 0.013 & 0.491 $\pm$ 0.023 
\\

\multirow{1}{*}{MAC~\cite{chen2025enhancing}}
& 0.666 $\pm$ 0.055 & 0.736 $\pm$ 0.035    
& 0.779 $\pm$ 0.069 & 0.731 $\pm$ 0.099
& 0.537 $\pm$ 0.011 & 0.544 $\pm$ 0.021 
& 0.530 $\pm$ 0.016 & 0.491 $\pm$ 0.020 
 \\

\multirow{1}{*}{AI-Hospital~\cite{fan2025ai}}
& 0.643 $\pm$ 0.038 & 0.767 $\pm$ 0.007   
& 0.810 $\pm$ 0.048 & 0.728 $\pm$ 0.061
& 0.547 $\pm$ 0.032 & 0.439 $\pm$ 0.026
& 0.581 $\pm$ 0.008 & 0.546 $\pm$ 0.016 
 \\

\rowcolor{gray!15} \multirow{1}{*}{CoMMa (Ours)}
&  \textbf{0.711 $\pm$ 0.011} & \textbf{0.795 $\pm$ 0.010} 
&  \textbf{0.837 $\pm$ 0.053} & \textbf{0.797 $\pm$ 0.038} 
&  \textbf{0.756 $\pm$ 0.171} & \textbf{0.666 $\pm$ 0.105} 
&  \textbf{0.711 $\pm$ 0.214} & \textbf{0.606 $\pm$ 0.138}  \\

\cmidrule(r){1-1} \cmidrule(r){2-3} \cmidrule(r){4-5} \cmidrule(r){6-7} \cmidrule(r){8-9}

\multicolumn{9}{l}{\textit{\small Qwen2.5-72B~\cite{qwen2_5}}} \\
\multirow{1}{*}{Single-Agent}
&  0.593 $\pm$ 0.072 & 0.650 $\pm$ 0.035    
&  0.813 $\pm$ 0.032 & 0.713 $\pm$ 0.069 
&  0.517 $\pm$ 0.015 & 0.543 $\pm$ 0.010 
&  0.502 $\pm$ 0.006 & 0.436 $\pm$ 0.004\\

\multirow{1}{*}{MDAgents~\cite{kim2024mdagents}}
& 0.602 $\pm$ 0.080 & 0.710 $\pm$ 0.024    
& 0.803 $\pm$ 0.034 & 0.750 $\pm$ 0.039 
& 0.517 $\pm$ 0.008 & 0.544 $\pm$ 0.020  
& 0.501 $\pm$ 0.023 & 0.437 $\pm$ 0.010 \\

\multirow{1}{*}{MAC~\cite{chen2025enhancing}}
& 0.689 $\pm$ 0.070 & 0.639 $\pm$ 0.044    
& 0.799 $\pm$ 0.050 & 0.761 $\pm$ 0.057
& 0.531 $\pm$ 0.009 & 0.543  $\pm$ 0.020 
& 0.525 $\pm$ 0.006 & 0.473 $\pm$ 0.004 
 \\

\multirow{1}{*}{AI-Hospital~\cite{fan2025ai}}
& 0.637 $\pm$ 0.100 & 0.727 $\pm$ 0.034   
& 0.827 $\pm$ 0.079 & 0.792 $\pm$ 0.090 
& 0.542 $\pm$ 0.015 & 0.544 $\pm$ 0.002 
& 0.517 $\pm$ 0.001 & 0.455 $\pm$ 0.010  
\\

\rowcolor{gray!15} \multirow{1}{*}{CoMMa (Ours)}
& \textbf{0.700 $\pm$ 0.014}& \textbf{0.750 $\pm$ 0.032}    
& \textbf{0.844 $\pm$ 0.020} & \textbf{0.802 $\pm$ 0.044} 
& \textbf{0.809 $\pm$ 0.090}& \textbf{0.666 $\pm$ 0.053} 
& \textbf{0.845 $\pm$ 0.107}& \textbf{0.576 $\pm$ 0.139}\\

\cmidrule(r){1-1} \cmidrule(r){2-3} \cmidrule(r){4-5} \cmidrule(r){6-7} \cmidrule(r){8-9}

\multicolumn{9}{l}{\textit{\small Gemma4-31B~\cite{google_gemma4_31b_it}}} \\
\multirow{1}{*}{Single-Agent}
& 0.625 $\pm$ 0.058 & 0.693 $\pm$ 0.047    
& 0.802 $\pm$ 0.063 & 0.815 $\pm$ 0.061 
& 0.556 $\pm$ 0.024 & 0.491 $\pm$ 0.045  
& 0.534 $\pm$ 0.017 & 0.491 $\pm$ 0.033\\

\multirow{1}{*}{MDAgents~\cite{kim2024mdagents}}
& 0.645 $\pm$ 0.014 & 0.718 $\pm$ 0.021   
& 0.826 $\pm$ 0.053 & 0.817 $\pm$ 0.064
& 0.590 $\pm$ 0.022 & 0.543 $\pm$ 0.018  
& 0.616 $\pm$ 0.030 & 0.564 $\pm$ 0.056 \\

\multirow{1}{*}{MAC~\cite{chen2025enhancing}}
& 0.645 $\pm$ 0.067 & 0.719 $\pm$ 0.039    
& 0.822 $\pm$ 0.044 & 0.827 $\pm$ 0.059
& 0.619 $\pm$ 0.040 & 0.597 $\pm$ 0.014 
& 0.597 $\pm$ 0.031 & 0.564 $\pm$ 0.017 \\

\multirow{1}{*}{AI-Hospital~\cite{fan2025ai}}
& 0.640 $\pm$ 0.028 & 0.747 $\pm$ 0.036   
& 0.791 $\pm$ 0.105 & 0.805 $\pm$ 0.081
& 0.616 $\pm$ 0.044 & 0.614 $\pm$ 0.019  
& 0.623 $\pm$ 0.014 & 0.583 $\pm$ 0.022 \\

\rowcolor{gray!15} \multirow{1}{*}{CoMMa (Ours)}
& \textbf{0.680 $\pm$ 0.028} &  \textbf{0.758 $\pm$ 0.027}    
& \textbf{0.841 $\pm$ 0.048} &  \textbf{0.830 $\pm$ 0.073} 
& \textbf{0.905 $\pm$ 0.083} &  \textbf{0.697 $\pm$ 0.105}
& \textbf{0.911 $\pm$ 0.077} &  \textbf{0.666 $\pm$ 0.139} \\

\bottomrule

\end{tabular}
}
\vspace{-3mm}
\end{table*}

\begin{table*}[t]
\centering
\caption{Ablation study on HCC Tumorboard treatment recommendation.}
\label{tab:ablation}
\resizebox{0.88\textwidth}{!}{%
\begin{tabular}{ccc cc cc}
\toprule
\multirow{2}{*}{Eqn.~(2)} & \multirow{2}{*}{$\mathcal{L}_\text{pg}$} & \multirow{2}{*}{$\mathcal{L}_\text{shap}$}
& \multicolumn{2}{c}{Llama3.1-70B} & \multicolumn{2}{c}{Qwen2.5-72B} \\
\cmidrule(lr){4-5} \cmidrule(lr){6-7}
& & & AUC & Acc & AUC & Acc \\
\midrule
\checkmark & \checkmark &            & $0.706 \pm 0.007$ & $0.688 \pm 0.068$ & $0.682 \pm 0.015$ & $0.738 \pm 0.004$ \\
\checkmark &            & \checkmark & $0.691 \pm 0.015$ & $0.659 \pm 0.042$ & $0.691 \pm 0.031$ & $0.728 \pm 0.015$ \\
           & \checkmark & \checkmark & $0.692 \pm 0.012$ & $0.741 \pm 0.056$ & $0.681 \pm 0.024$ & $0.722 \pm 0.046$ \\
\checkmark & \checkmark & \checkmark
& \textbf{0.711 $\pm$ 0.011}
& \textbf{0.795 $\pm$ 0.010}
& \textbf{0.700 $\pm$ 0.014}
& \textbf{0.750 $\pm$ 0.032} \\
\midrule
\multicolumn{3}{l}{1.~Replace learned weights with uniform dist.}
& $0.693 \pm 0.008$ & $0.683 \pm 0.072$ & $0.625 \pm 0.023$ & $0.661 \pm 0.055$ \\
\multicolumn{3}{l}{2.~Sampled-coalition Shapley approximation.}
& $0.689 \pm 0.011$ & $0.717 \pm 0.049$ & $0.698 \pm 0.013$ & $0.725 \pm 0.029$ \\
\multicolumn{3}{l}{3.~Use centralized input.}
& $0.691 \pm 0.006$ & $0.734 \pm 0.055$ & $0.692 \pm 0.008$ & $0.613 \pm 0.086$ \\
\bottomrule
\end{tabular}%
}
\end{table*}

\subsection{Experimental Analysis}
\label{sec:analysis}
\yichen{To further investigate the behavior of CoMMa beyond the quantitative performance improvements reported above, we conduct additional analyses to better understand how the proposed framework achieves its gains, whether the learned contribution patterns are clinically meaningful, and how well the framework generalizes across different model settings and medical foundation backbones. Specifically, the following analyses are organized around three key questions.}

\textbf{Question 1: How does each component of CoMMa contribute to performance?} \yichen{Table~\ref{tab:ablation} reports ablation results on HCC treatment recommendation. Removing either contribution-learning objective degrades performance, showing that $\mathcal{L}_{\mathrm{pg}}$ and $\mathcal{L}_{\mathrm{shap}}$ provide complementary supervision for the contribution gate. 
We further analyze the effect of contribution-aware aggregation. Replacing the learned contribution weights with uniform averaging degrades both AUC and Acc, showing that patient-specific agent weighting is important for effective multi-agent collaboration. In addition, replacing exact Shapley computation with Monte Carlo approximation leads to lower performance, suggesting that accurate coalition-aware contribution estimation improves the stability of contribution learning. Finally, using centralized input degrades performance, indicating the benefit of decentralized data streams. Overall, the ablation results demonstrate that both contribution-aware weighting and Shapley-based regularization are important components of the proposed CoMMa algorithm. }


\begin{figure*}[t]
\centering
\includegraphics[width=1\linewidth]{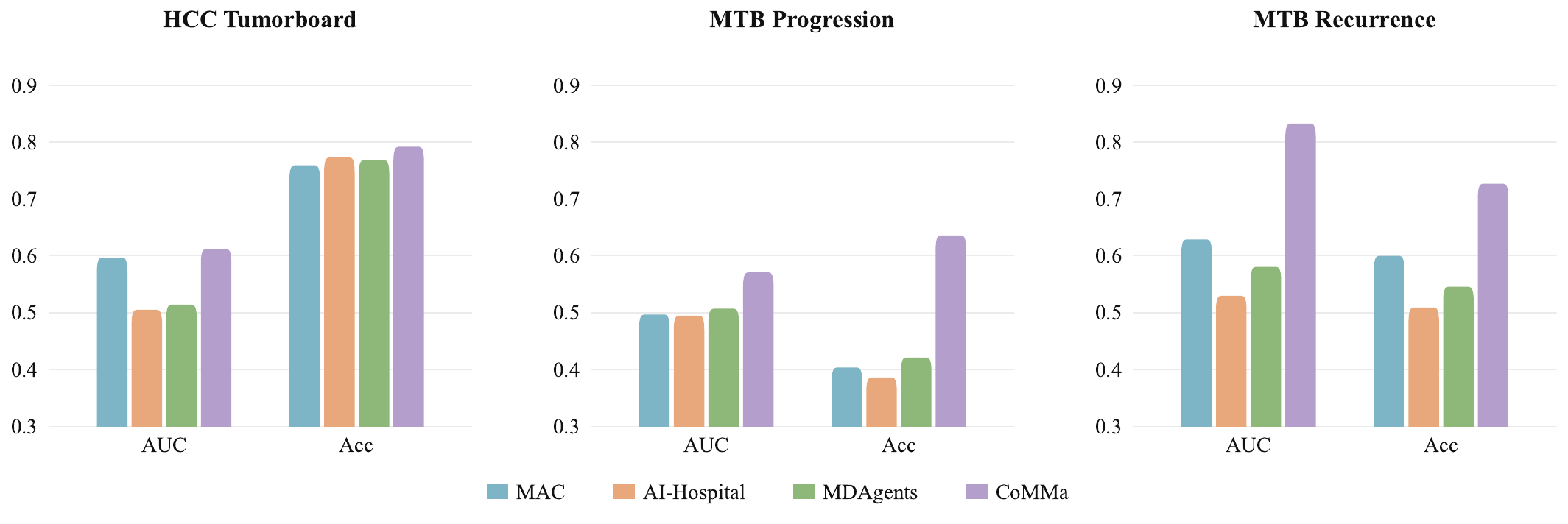}
\caption{Comparison of AUC/Acc across different methods using the medical-related Meditron-70B backbone~\cite{meditron2023} on multiple clinical evaluation tasks.}
\label{fig:medirtron}
\vspace{-3mm}
\end{figure*}

\begin{wraptable}{r}{0.48\textwidth}
\vspace{-20pt}
\centering
\small
\caption{Fine-tuned baselines (mean of 3 seeds).}
\label{tab:finetune_baseline}
\vspace{4pt}
\resizebox{\linewidth}{!}{%
\begin{tabular}{lcccccc}
\toprule
\multirow{2}{*}{\textbf{HCC Tumorboard}} & \multicolumn{2}{c}{\textbf{Llama}} & \multicolumn{2}{c}{\textbf{Qwen}} & \multicolumn{2}{c}{\textbf{Gemma}} \\
\cmidrule(lr){2-3} \cmidrule(lr){4-5} \cmidrule(lr){6-7}
& AUC & Acc & AUC & Acc & AUC & Acc \\
\midrule
AI-Hospital-FT & 0.676 & 0.753 & 0.684 & 0.738 & 0.607 & 0.751 \\
\textbf{CoMMa}  & \textbf{0.711} & \textbf{0.795} & \textbf{0.700} & \textbf{0.750} & \textbf{0.680} & \textbf{0.758} \\
\bottomrule
\end{tabular}%
}
\vspace{-10pt}
\end{wraptable}
\textbf{Question 2: What if other multi-agent methods also adopt finetuning?} To examine whether CoMMa's improvement is simply due to supervised adaptation, we introduce fine-tuned variants of existing SOTA multi-agent baselines, AI-Hospital-FT. All fine-tuned baselines use the same backbone, data split, prediction labels, training objective, and evaluation protocol as CoMMa. As shown in Table~\ref{tab:finetune_baseline}, finetuning improves these baselines over their prompt-only versions, but CoMMa still achieves the best overall performance. This indicates that CoMMa's gains are not merely a result of finetuning, but are also attributable to decentralized evidence specialization, latent cross-agent communication, and contribution-aware aggregation. Other finetuned baselines are in the~\ref{app:finetuned_baselines}.

\textbf{Question 3: Does CoMMa remain effective with medical foundation models?}
As shown in Fig.~\ref{fig:medirtron}, we evaluate CoMMa on Meditron-70B~\cite{meditron2023}, where it consistently delivers improvements over strong domain-specific LLMs. CoMMa achieves higher AUC and accuracy across all tasks, demonstrating its effectiveness even when built on top of medical-specialized backbones. These results indicate that CoMMa provides complementary gains beyond model pretraining and remains robust across different backbone choices and diverse clinical evaluation settings. 

\begin{figure*}[t]
  \centering
  \includegraphics[width=1\linewidth]{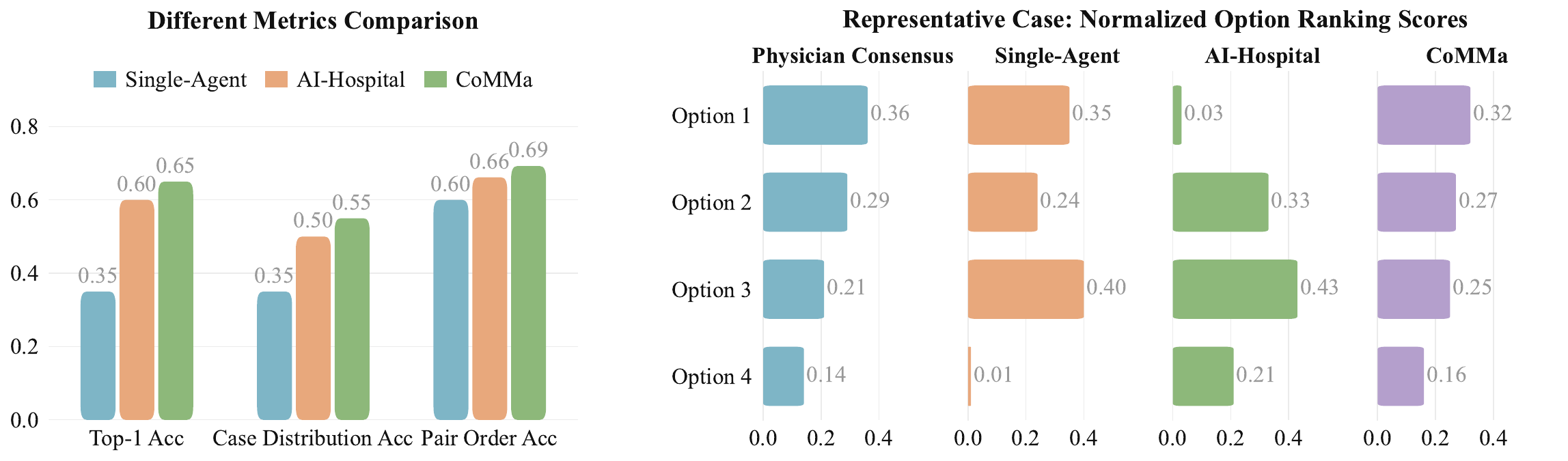}
   \caption{
Performance and representative case analysis on \textsc{TumorBoardRank} using Gemma4-31B.
Left: overall comparison.
Right: normalized option-ranking scores for a representative case, with physician consensus as the reference distribution.
}
   \vspace{-2mm}
   \label{fig:case_charts}
\end{figure*}

\begin{figure*}[t]
  \centering
  \includegraphics[width=1\linewidth]{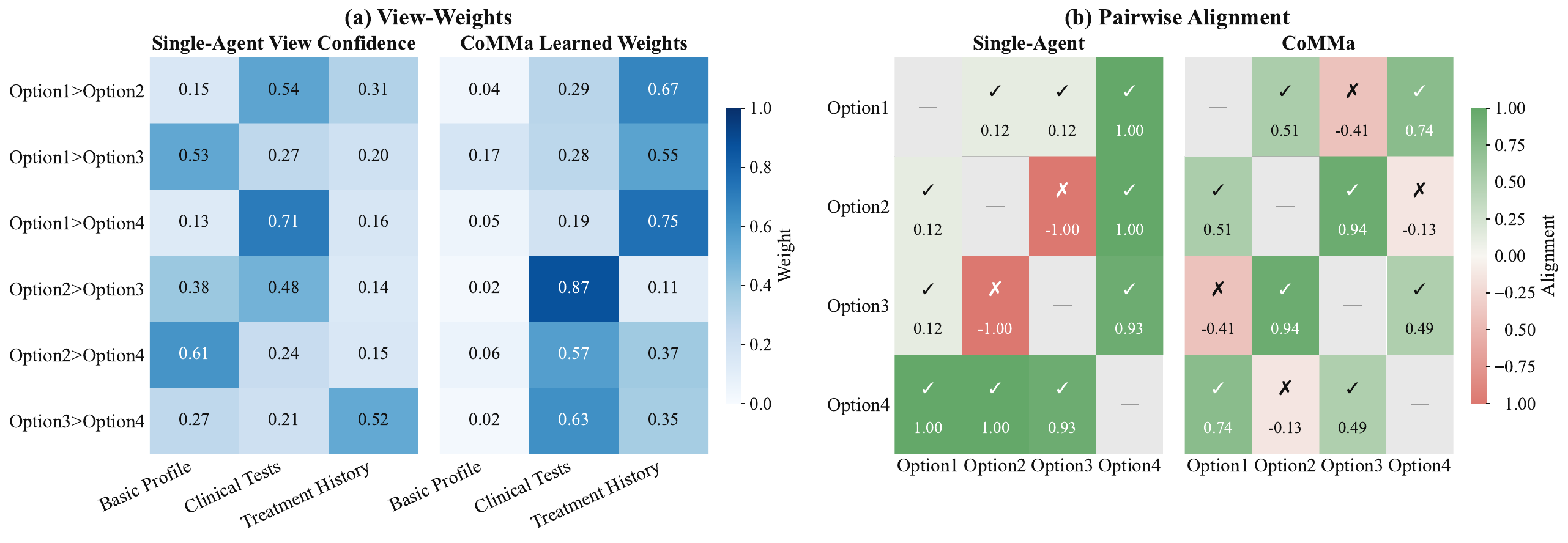}
   \caption{
View-weight and pairwise-alignment analysis for a representative \textsc{TumorBoardRank} case using Gemma4-31B.
(a) View-level confidence and learned CoMMa weights across evidence streams.
(b) Pairwise alignment with physician-consensus option ordering, where sign indicates agreement and magnitude indicates confidence.
}
   \vspace{-2mm}
   \label{fig:alignment_weights}
\end{figure*}

\textbf{Question 4: Does CoMMa provide clinically meaningful contribution attribution?}
We conduct this analysis on \textsc{TumorBoardRank}, an externally annotated tumor board option-ranking dataset with physician-consensus preferences. As shown in Fig.~\ref{fig:case_charts}, we
compare CoMMa with Single-Agent and AI-Hospital, the strongest overall baseline. Using Gemma4-31B, CoMMa improves Top-1 Acc, Case Distribution Acc, and Pair Order Acc over both baselines, suggesting
better recovery of physician-consensus rankings. In the representative case, CoMMa produces a normalized ranking distribution that better follows physician consensus, while the baselines over-weight
lower-ranked alternatives.
Fig.~\ref{fig:alignment_weights} further analyzes the same case. Single-Agent view confidence is computed from frozen Gemma4-31B view-level logit margins, whereas CoMMa weights are learned by the
fusion module. CoMMa assigns larger weights to clinically informative streams, such as clinical tests and treatment history, and down-weights the less informative basic profile view, indicating
learned evidence reweighting beyond raw view confidence. Although CoMMa makes one more pairwise error than the view-level Gemma4 baseline, this error has low confidence, whereas the baseline makes a
high-confidence error. This explains why CoMMa recovers the correct case-level distribution in Fig.~\ref{fig:case_charts}, while the baseline assigns excessive mass to a lower-ranked option.

\section{Conclusion and  Limitation}
In this work, we propose CoMMa, a contribution-aware medical multi-agent framework that replaces prompt-based role interaction with decentralized specialist modeling and structured contribution learning. By combining evidence-specific specialist agents, deterministic cross-agent collaboration, and Shapley-regularized contribution aggregation, CoMMa provides improved predictive performance and explicit agent-level attribution. Experiments across multiple oncology benchmarks demonstrate consistent gains over single-agent and existing multi-agent baselines under different open-source LLM backbones. In addition, the decentralized design of CoMMa better supports privacy-sensitive and resource-constrained clinical deployment settings by avoiding full-context sharing and expensive cloud-based inference. Future work will extend CoMMa beyond oncology benchmarks to broader clinical domains and more heterogeneous healthcare settings. We also plan to investigate finer-grained contribution estimation and representation disentanglement to better model complex interactions among correlated clinical evidence streams. Finally, extending contribution-aware multi-agent reasoning to open-ended clinical decision-making scenarios, including ambiguous outcomes and free-form clinical reasoning, represents an important direction for future research.

\clearpage
\bibliographystyle{plain}
\bibliography{ref}

\clearpage
\appendix
\section{Appendix}


\subsection{Dataset Details.} \label{app:data}

\subsubsection{HCC Tumorboard}

\kailong{
The HCC Tumorboard dataset comprises 116 patients diagnosed with primary Hepatocellular Carcinoma (HCC) at a large tertiary academic hospital in North America. The study was conducted under institutional IRB approval. The cohort includes adult patients with confirmed hepatocellular carcinoma, no concurrent malignancies, and multidisciplinary tumor board review during the study period. Each case was reviewed by a multidisciplinary tumor board to establish the consensus treatment recommendation, which serves as the ground-truth label. Cases without recommendations in all treatment categories
were excluded. We adopt a 75/25\% train-test split.
The ground-truth label is defined over eight discrete treatment options: (A) transcatheter arterial chemoembolization (TACE), (B) ablation, primarily microwave ablation (MWA), (C) radioembolization/
selective internal radiation therapy (SIRT), (D) radiotherapy, (E) surgical resection, (F) liver transplant, (G) chemotherapy, and (H) palliative care. Each instance contains 24 multimodal clinical
variables represented in textual format, extracted from longitudinal clinical records and MRI reports. An exemplar case illustrating the HCC Tumorboard data construction is provided
in~\cref{tab_hcc_data}.
}

\begin{table}[ht]
  \caption{HCC Tumorboard Data Formation.}
  \centering
  \resizebox{\linewidth}{!}{
  \renewcommand{\arraystretch}{1}
  \begin{tabular}{lp{4.8cm}p{4.8cm}p{4.8cm}l}
      \toprule
       & \multicolumn{4}{c}{\textbf{Decentralized Data}} \\
      \cmidrule(lr){2-5}
       & $\mathbf{x}_1$ & $\mathbf{x}_2$ & $\mathbf{x}_3$ & $\mathbf{x}_4$ \\
       & MRI imaging features & OPTN$^\dagger$ information & Performance score & Laboratory data \\
      \cmidrule(l){1-1} \cmidrule(l){2-2}  \cmidrule(l){3-3}  \cmidrule(l){4-4} \cmidrule(l){5-5}
      Context example
      & \begin{tabular}[t]{@{}l@{}}
      Number of tumors: 4.0 \\
      Size of largest tumor (cm): 2.6 \\
      Location largest (segment): 8 \\
      Vascular invasion: 0 \\
      Extrahepatic spread: 0 \\
      Thrombus (portal vein): 0 \\
       \vdots \\
      \end{tabular}
      & \begin{tabular}[t]{@{}l@{}}
      Number of OPTN 5A tumors: 0 \\
      Number of OPTN 5B tumors: 2 \\
      Number of OPTN 5X tumors: 0 \\
      Number of OPTN 5T tumors: 0
      \end{tabular}
      & \begin{tabular}[t]{@{}l@{}}
      Child-Pugh score: A \\
      ECOG$^\ddagger$ score: 1 \\
      MELD$^\S$ score: 13.0
      \end{tabular}
      & \begin{tabular}[t]{@{}l@{}}
      Encephalopathy: 0 \\
      Ascites: 0 \\
      Bilirubin: 0.6 \\
      Albumin: 4.8 \\
      INR: 1.1 \\
      AFP: 5.0 \\
      \vdots
      \end{tabular} \\
      \bottomrule
  \multicolumn{5}{l}{$^\dagger$OPTN: Organ Procurement \& Transplantation Network, $^\ddagger$ECOG: Eastern Cooperative Oncology Group Performance, $^\S$MELD: Model for End-stage Liver Disease}
  \end{tabular}
  }
  \label{tab_hcc_data}
\end{table}

\subsubsection{MTBBench}
MTBBench~\cite{vasilev2025mtbbench} is an agentic benchmark designed to evaluate molecular tumor board-style reasoning that requires integrating multimodal evidence and longitudinal patient context.
It contains 66 patient cases and 573 expert-validated question-answer pairs across two tracks. The multimodal track includes 26 head and neck cancer cases, with approximately 40 modality-specific files per patient, including H\&E slides, IHC images, and hematology reports, yielding 390 questions. The longitudinal track includes 40 cases with approximately five structured files per patient
and 183 questions focused on time-dependent outcomes, including recurrence and progression.
In our experiments, we focus on prognosis prediction tasks for tumor recurrence and progression. Each sample is formulated as a paired clinical question and binary endpoint label. We adopt an 80/20 \%
train-test split for MTBBench. An exemplar case illustrating the MTBBench data construction is provided in~\cref{tab_mtb_data}.

\begin{table}[ht]
\caption{MTBBench Data Formation.}
\centering
\resizebox{\linewidth}{!}{
\renewcommand{\arraystretch}{1}
\begin{tabular}{lllll}
    \toprule
     & \multicolumn{4}{c}{\textbf{Decentralized Data}} \\
    \cmidrule(lr){2-5}
     & $\mathbf{x}_1$ & $\mathbf{x}_2$ & $\mathbf{x}_3$ & $\mathbf{x}_4$ \\
     & Specimen metadata & Somatic mutations & {Copy number alterations} &  {Structural variants} \\
    \cmidrule(l){1-1} \cmidrule(l){2-2}  \cmidrule(l){3-3}  \cmidrule(l){4-4} \cmidrule(l){5-5}
    
    Context example
    & \begin{tabular}[t]{@{}l@{}}
    Colon Adenocarcinoma \\
    Primary (Sigmoid Colon) \\
    Tumor Purity: 20.0\% \\
    MSI Type: Stable \\
    \vdots
    \end{tabular}
    & \begin{tabular}[t]{@{}l@{}}
    TP53 p.C242F \\
    FANCA p.V750M \\
    CDK12 p.S238$*$\\
    \vdots \\
    \end{tabular}
    & \begin{tabular}[t]{@{}l@{}}
    ERBB2 gain (Value: 2.0)\% \\
    CDK12 gain (Value: 2.0) \\
    \vdots \\ \\
    \end{tabular}
    & \begin{tabular}[t]{@{}l@{}}
    CTNNB1 Splice Site\\ 
    deletion (c.18$\_$242-37del) \\ 
    \vdots \\ \\
    \end{tabular}
    \\
    \bottomrule
    
\end{tabular}
}
\label{tab_mtb_data}
\end{table}

\subsubsection{LungCure}
LungCure~\cite{hao2026lungcure} is a lung cancer staging benchmark designed to evaluate clinical reasoning from longitudinal OCR-derived patient records. It contains 999 patients and We adopt an 80/20 \% train-test split. In our experiments, we focus on six binary staging indicators spanning primary tumor extent (T), regional nodal involvement (N), and distant metastasis (M): T2a, T1c, N0, N2, M0,
and M1c. Each sample is formulated as a patient record paired with a binary TNM staging label. We partition each record into TNM-oriented evidence streams, including primary tumor evidence, nodal
evidence, metastasis evidence, and clinical history. An exemplar case illustrating the LungCure data construction is provided in~\cref{tab_lungcure_data}.

\begin{table}[ht]
  \caption{LungCure Data Formation.}
  \centering
  \resizebox{\linewidth}{!}{
  \renewcommand{\arraystretch}{1}
  \begin{tabular}{lllll}
      \toprule
       & \multicolumn{4}{c}{\textbf{Decentralized Data}} \\
      \cmidrule(lr){2-5}
       & $\mathbf{x}_1$ & $\mathbf{x}_2$ & $\mathbf{x}_3$ & $\mathbf{x}_4$ \\
       & Primary tumor evidence & Regional nodal evidence & Distant metastasis evidence & Clinical history \\
      \cmidrule(l){1-1} \cmidrule(l){2-2} \cmidrule(l){3-3} \cmidrule(l){4-4} \cmidrule(l){5-5}

      Context example
  & \begin{tabular}[t]{@{}l@{}}
  Right upper-lobe solid nodule \\
  Size: 3.1$\times$3.3$\times$2.7 cm \\
  SUVmax: 8.1 \\
  Spiculation and pleural traction \\
  \vdots
  \end{tabular}
  & \begin{tabular}[t]{@{}l@{}}
  Right hilar lymph nodes enlarged \\
  Right paratracheal lymph nodes \\
  Increased FDG uptake \\
  SUVmax: 5.9 \\
  \vdots
  \end{tabular}
  & \begin{tabular}[t]{@{}l@{}}
  No definite distant metastasis \\
  No abnormal liver uptake \\
  No pelvic or retroperitoneal nodes \\
  T4 bone island, favored benign \\
  \vdots
  \end{tabular}
  & \begin{tabular}[t]{@{}l@{}}
  Smoking history over 30 years \\
  Family history of ovarian cancer \\
  PET/CT for systemic assessment \\
  Thoracic surgery outpatient referral \\
  \vdots
  \end{tabular}
      \\
      \bottomrule

  \end{tabular}
  }
  \label{tab_lungcure_data}
\end{table}

\subsubsection{\textsc{TumorBoardRank}}
\textsc{TumorBoardRank} is an externally annotated tumor board option-ranking dataset for evaluating cross-domain clinical decision ranking and interpretability. It contains 93 tumor board cases
after filtering single-option cases, yielding 276 pairwise option comparisons. Each case consists of a refined clinical case summary, physician-consensus preferences over candidate management
options, and the corresponding option descriptions. We use a case-level temporal split, with 73 cases for training and 20 cases for testing. For CoMMa, each case is partitioned into three evidence
streams: basic profile, clinical tests, and treatment history. An exemplar case illustrating the \textsc{TumorBoardRank} data construction is provided in~\cref{tab_tumorboardrank_data}.

\begin{table}[ht]
  \caption{\textsc{TumorBoardRank} Data Formation.}
  \centering
  \resizebox{\linewidth}{!}{
  \renewcommand{\arraystretch}{1}
  \begin{tabular}{llll}
      \toprule
       & \multicolumn{3}{c}{\textbf{Decentralized Data}} \\
      \cmidrule(lr){2-4}
       & $\mathbf{x}_1$ & $\mathbf{x}_2$ & $\mathbf{x}_3$ \\
       & Basic profile & Clinical tests & Treatment history \\
      \cmidrule(l){1-1} \cmidrule(l){2-2} \cmidrule(l){3-3} \cmidrule(l){4-4}

      Context example
      & \begin{tabular}[t]{@{}l@{}}
      37-year-old man \\
      Laryngeal squamous cell carcinoma \\
      ECOG performance status: 0 \\
      No chronic high-grade toxicity \\
      \vdots
      \end{tabular}
      & \begin{tabular}[t]{@{}l@{}}
      Isolated nodal recurrence \\
      Salvage neck dissection \\
      1 of 10 lymph nodes positive \\
      Extranodal extension present \\
      \vdots
      \end{tabular}
      & \begin{tabular}[t]{@{}l@{}}
      Prior definitive chemoradiation \\
      69.96 Gy in 33 fractions \\
      56.1 Gy elective nodal irradiation \\
      Cisplatin cumulative dose: 200 mg/m$^2$ \\
      \vdots
      \end{tabular}
      \\
      \bottomrule

  \end{tabular}
  }
  \label{tab_tumorboardrank_data}
\end{table}

\subsection{Detailed Shapley Computation.} \label{app:sha}


\subsubsection{Comparison to Advantage Vector} Unlike the advantage vector $\mathbf{A}_i$, which only considers the marginal impact of removing agent $i$ from the full coalitions $\mathcal{N}$, Shapley value evaluates the contribution of agent $i$ across a diverse set of sub-coalitions $S \subseteq \mathcal{N} \setminus \{i\}$. of varying cardinalities. Consequently, the resulting Shapley matrix $\mathbf{\Phi}$ captures an expected marginal contribution that explicitly accounts for inter-agent interactions and dependencies, rather than being dominated by sample-specific or coalition-dependent fluctuations.

\subsection{Comparison on Computational Cost.} \label{app:comp}

We compare computational complexity and cost across trainable baselines, single-agent in-context learning, role-based multi-agent systems, and our locally fine-tunable multi-agent approach in~\cref{tab_compute}. Latency and inference token usage are averaged over 30 samples from the HCC Tumorboard dataset. Our results show that CoMMa is a cost-effective method, achieving lower latency under a free-inference setting while delivering superior performance.

\begin{table}[ht]
\caption{Computational Complexity and cost Comparison on HCC Tumorboard Dataset.}
\centering
\resizebox{\linewidth}{!}{  
\begin{tabular}{lcccc}
\toprule
\multirow{2}{*}{\bf{Methods}}  & \bf{Trainable Baselines} & \bf{Single-Agent In-context Learning} & \bf{Role-based Multi-Agents} & \bf{Locally Fine-tunable Multi-Agents} \\ 
 \cmidrule(l){2-2}  \cmidrule(l){3-3}  \cmidrule(l){4-4} \cmidrule(l){5-5} 
& {CatBoost} & GPT 4.1 & {MDAgents} &  {CoMMa}  \\    
\cmidrule(l){1-1} \cmidrule(l){2-2}  \cmidrule(l){3-3}  \cmidrule(l){4-4} \cmidrule(l){5-5} 
Computational Complexity & 0.048 MFLOPs & Unreleased & Unreleased & 280 TFLOPs \\
Total Trainable Parameters & 0.512 M & None & None & 16.8 M \\
Latency (per input) & 0.116 ms  & 4.2 s & 79.2 s & 1.1 s \\
Inference Tokens (chargeable) & None & 20,878 & 23,254 & None \\

\bottomrule 
\end{tabular}
}
\label{tab_compute}
\end{table}

\subsection{Fine-Tuned Multi-Agent Baselines}
  \label{app:finetuned_baselines}

To assess whether CoMMa's gains can be explained solely by supervised fine-tuning, we construct fine-tuned variants of representative multi-agent baselines. These baselines use the same LLM
backbone, train-test split, prediction labels, supervision objective, and evaluation protocol as CoMMa. The main paper reports AI-Hospital-FT, while additional
fine-tuned baselines are summarized in Tabel~\ref{tab:finetune_baseline_appendix}.
For each baseline, we preserve its original multi-agent organization whenever possible and add a supervised prediction head trained on the task-specific labels. The frozen LLM encoder is used to
obtain case-level or agent-level representations, and the trainable head is optimized with the same binary cross-entropy objective used by CoMMa. This setup controls for the effect of task-specific
supervision while retaining each baseline's original agent structure.

\begin{table}[t]
\centering
\small
\caption{Comparison with fine-tuned multi-agent baselines on HCC Tumorboard. Results are averaged over 3 random seeds.}
\label{tab:finetune_baseline_appendix}
\begin{tabular}{lcccccc}
\toprule
\multirow{2}{*}{\textbf{HCC}} & \multicolumn{2}{c}{\textbf{Llama}} & \multicolumn{2}{c}{\textbf{Qwen}} & \multicolumn{2}{c}{\textbf{Gemma}} \\
\cmidrule(lr){2-3} \cmidrule(lr){4-5} \cmidrule(lr){6-7}
& AUC & Acc & AUC & Acc & AUC & Acc \\
\midrule
MAC            & 0.656 & 0.780 & 0.690 & 0.738 & 0.650  & 0.739 \\
MDAgents       & 0.676 & 0.776 & 0.685 & 0.741 & 0.626  & 0.731 \\
AI-Hospital    & 0.676 & 0.753 & 0.684 & 0.738 & 0.607  & 0.751 \\
\midrule
\textbf{CoMMa} & \textbf{0.711} & \textbf{0.795} & \textbf{0.700} & \textbf{0.750} & \textbf{0.680} & \textbf{0.758} \\
\bottomrule
\end{tabular}
\end{table}


\subsection{Detailed Class-wise Performance Comparison on HCC Treatments.} \label{app:hcc_cls}

\subsection{Additional Response Comparison.} \label{app:res}

\subsubsection{HCC Tumorboard}

\begin{figure*}[!ht]
  \centering
  \includegraphics[width=1\linewidth]{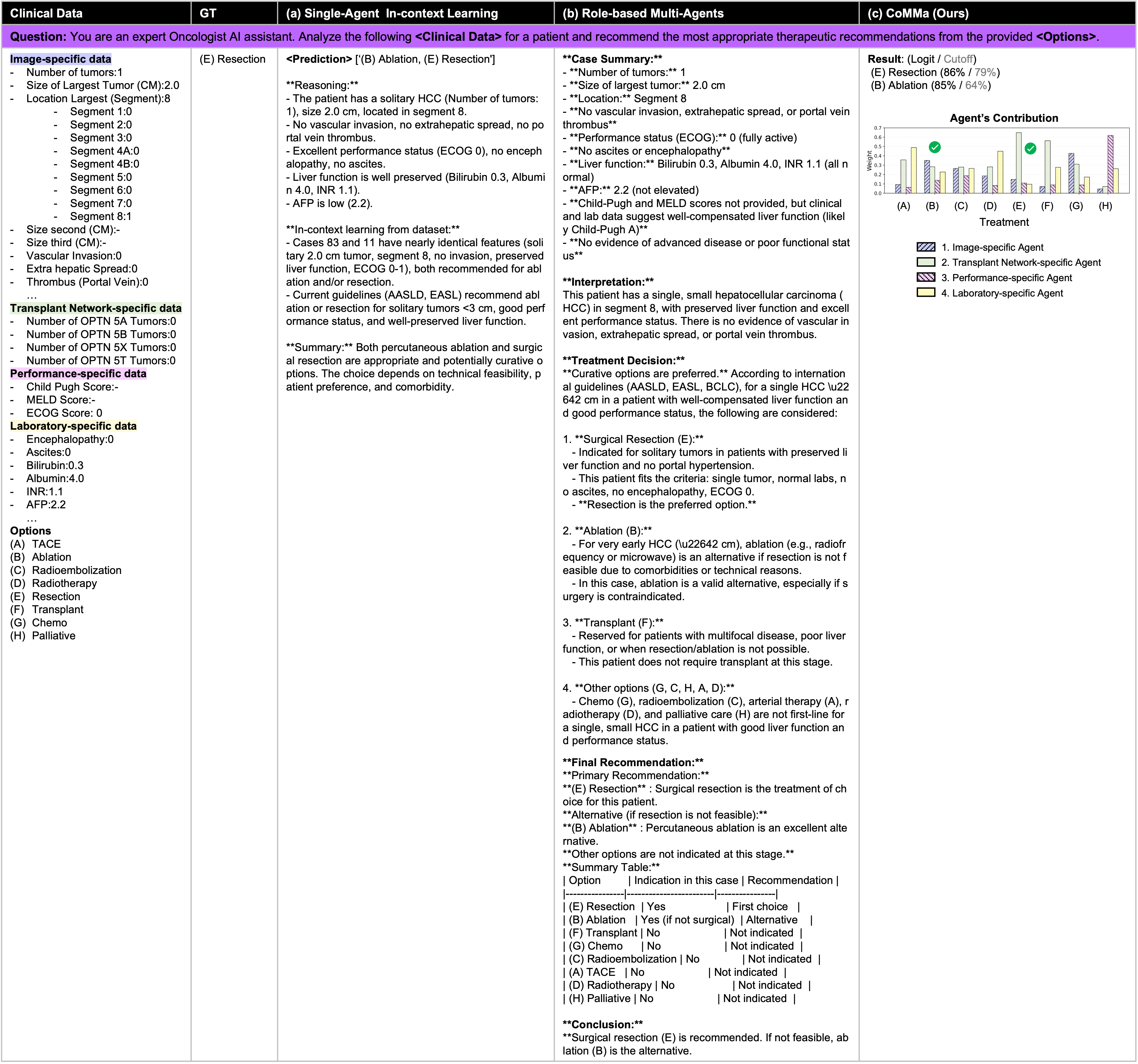}
   \caption{Additional agentic framework response comparison on HCC Tumorboard dataset - \#1. (a) Single-agent in-context learning (GPT-4.1), (b) Role-based multi-agent framework (MDAgents-Advanced), and (c) CoMMa. For the role-based multi-agent, intermediate discussions are omitted, and only the final consensus is visualized. For CoMMa, class-wise logits and cutoff thresholds are transformed into percentiles.}
   \vspace{-10pt}
   \label{fig:res2}
\end{figure*}

\begin{figure*}[!ht]
  \centering
  \includegraphics[width=1\linewidth]{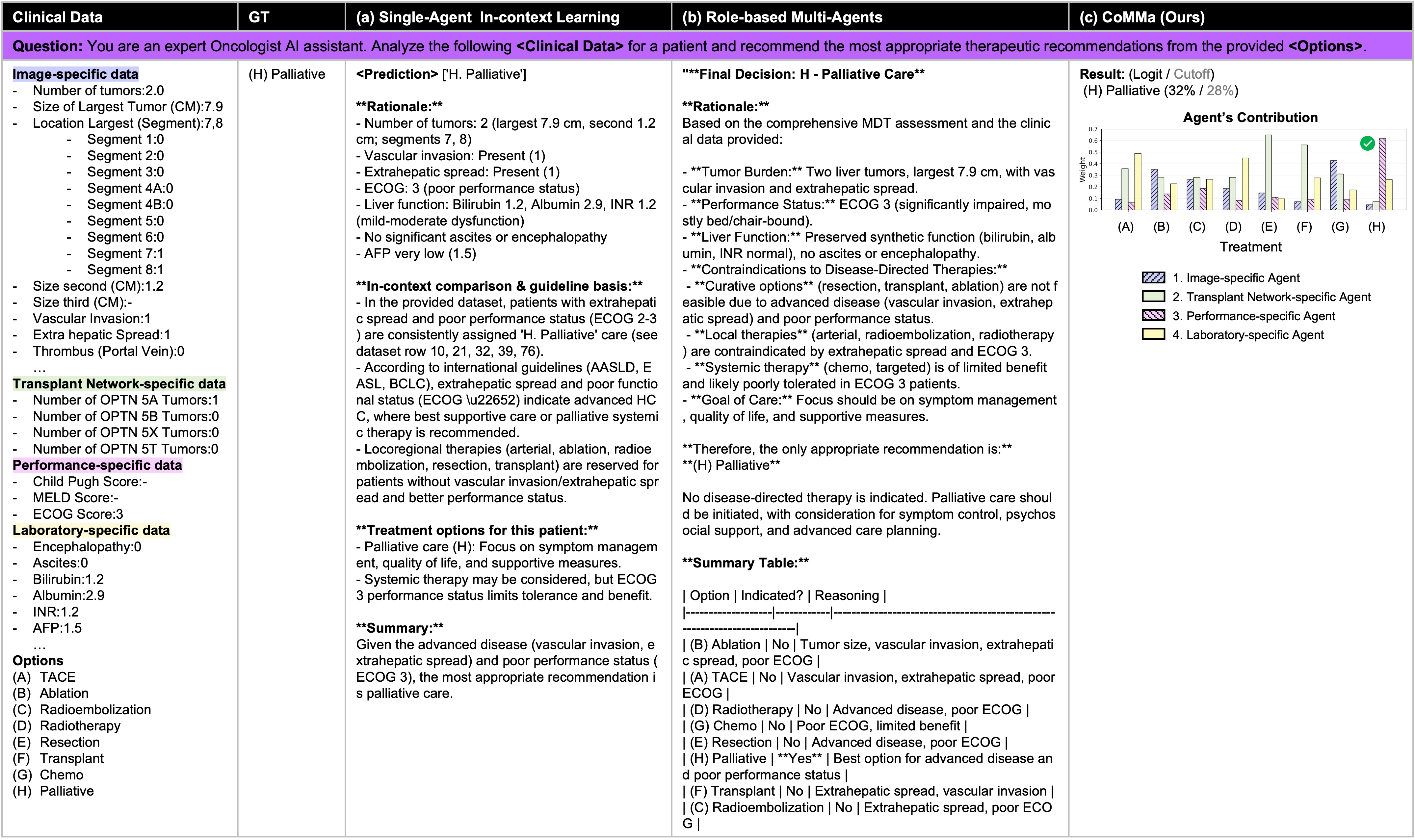}
   \caption{Additional agentic framework response comparison on  HCC Tumorboard dataset - \#2.(a) Single-agent in-context learning (GPT-4.1), (b) Role-based multi-agent framework (MDAgents-Advanced), and (c) CoMMa. For the role-based multi-agent, intermediate discussions are omitted, and only the final consensus is visualized. For CoMMa, class-wise logits and cutoff thresholds are transformed into percentiles.}
   \vspace{-10pt}
   \label{fig:res3}
\end{figure*}

We further provide agentic framework response comparison on HCC Tumorboard dataset in~\cref{fig:res2}  and~\cref{fig:res3}. Whereas single-agent in-context learning and role-based multi-agent methods often yield verbose yet clinically inconsistent rationales, CoMMa provides quantitative attribution of agent contributions within a structured decision framework, resulting in superior predictive performance.

\subsubsection{MTBBench}

\begin{figure*}[!ht]
  \centering
  \includegraphics[width=1\linewidth]{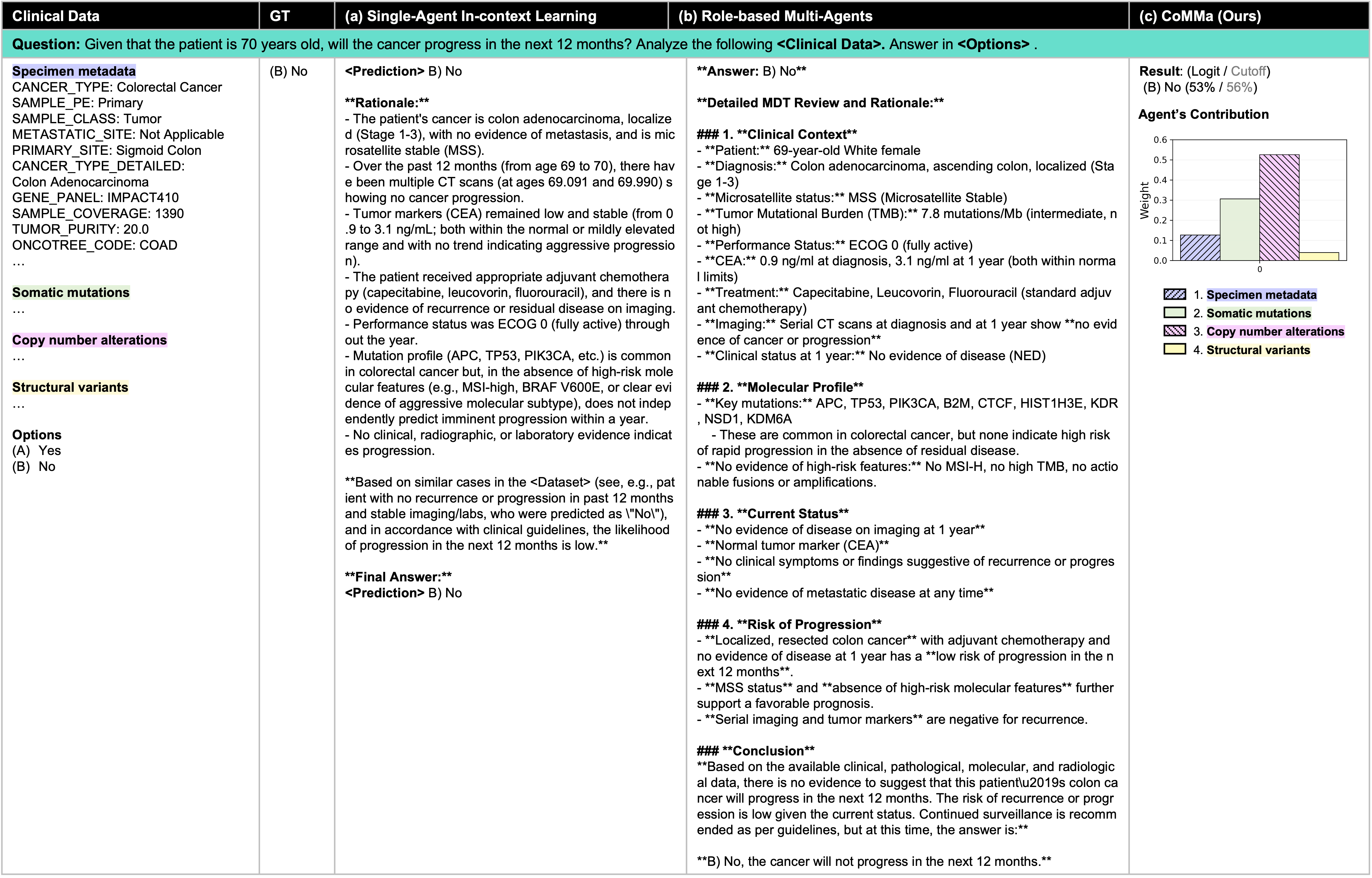}
   \caption{Additional agentic framework response comparison on  MTBBench - tumor progress question. (a) Single-agent in-context learning (GPT-4.1), (b) Role-based multi-agent framework (MDAgents-Advanced), and (c) CoMMa. For the role-based multi-agent, intermediate discussions are omitted, and only the final consensus is visualized. For CoMMa, class-wise logits and cutoff thresholds are transformed into percentiles.}
   \vspace{-10pt}
   \label{fig:mtb_1_0040923_3}
\end{figure*}

\begin{figure*}[!ht]
  \centering
  \includegraphics[width=1\linewidth]{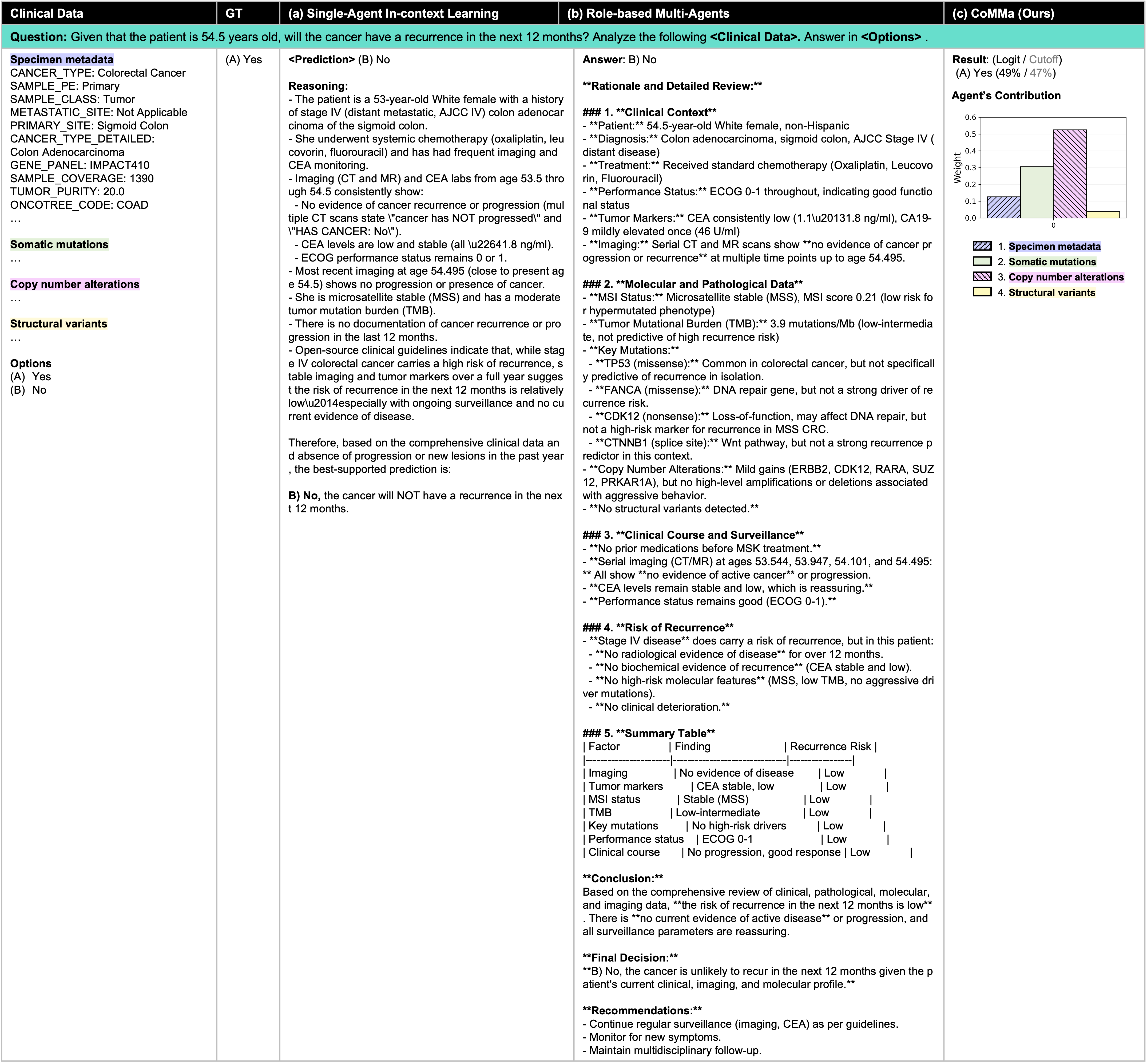}
   \caption{Additional agentic framework response comparison on  MTBBench - tumor recurrence question. (a) Single-agent in-context learning (GPT-4.1), (b) Role-based multi-agent framework (MDAgents-Advanced), and (c) CoMMa. For the role-based multi-agent, intermediate discussions are omitted, and only the final consensus is visualized. For CoMMa, class-wise logits and cutoff thresholds are transformed into percentiles.}
   \vspace{-10pt}
   \label{fig:mtb_1_0006687_2}
\end{figure*}

We provide a comparison of agentic framework responses on the MTBbench dataset. Specifically, we present a representative sample from each tumor progression tasks in~\cref{fig:mtb_1_0040923_3} and tumor recurrence prediction tasks in~ \cref{fig:mtb_1_0006687_2}. We observe that despite the extensive reasoning produced by both single-agent in-context learning and role-based multi-agent frameworks, their predictive accuracy remains significantly inferior to our proposed method.

\clearpage


\end{document}